\documentclass[lettersize,journal]{IEEEtran}
\usepackage{amsmath,amsfonts}
\usepackage{algorithmic}
\usepackage{array}
\usepackage[caption=false,font=normalsize,labelfont=sf,textfont=sf]{subfig}
\usepackage{textcomp}
\usepackage{stfloats}
\usepackage{url}
\usepackage{verbatim}
\usepackage{graphicx}
\usepackage{cite}
\usepackage{textcomp}
\usepackage{multirow}
\usepackage{bm}
\usepackage{bbm}
\usepackage{xcolor}
\usepackage{booktabs}
\usepackage[normalem]{ulem}
\useunder{\uline}{\ul}{}
\usepackage[ruled,linesnumbered]{algorithm2e}
\begin{document}

\title{DredRL: Dual Representation Enhanced Diffusion Policy for Flexible Job Shop Scheduling}

\author{Yunyi Zhao, \IEEEmembership{Student Member, IEEE}, Wei Zhang, \IEEEmembership{Member, IEEE}, Cheng Xiang, \IEEEmembership{Member, IEEE}, Vadakkepat Prahlad, \IEEEmembership{Member, IEEE} and Zuming Liu, \IEEEmembership{Member, IEEE}
\thanks{Manuscript received November 14, 2026; revised XXX, 2026; accepted XXX, 2026. (\textit{Corresponding author: Wei Zhang})}
\thanks{Yunyi Zhao is with both the Information and Communications Technology Cluster, Singapore Institute of Technology, Singapore 138683, and the Department of Electrical and Computer Engineering, National University of Singapore, Singapore 119077 (e-mail: yunyi.zhao@singaporetech.edu.sg, zhaoyunyi@u.nus.edu).}
\thanks{Wei Zhang is with the Information and Communications Technology Cluster, Singapore Institute of Technology, Singapore 138683 (e-mail: wei.zhang@singaporetech.edu.sg).}
\thanks{Cheng Xiang is with the School of Information and Communication Engineering, Dalian University of Technology, Dalian, Liaoning Province, China, 116024 (e-mail: xiangcheng@dlut.edu.cn).}
\thanks{Vadakkepat Prahlad is with the Department of Electrical and Computer Engineering, National University of Singapore, Singapore 119077 (e-mail: prahlad@nus.edu.sg).}
\thanks{Zuming Liu is with the College of Smart Energy, Shanghai Jiao Tong University, Shanghai, China, 200240 (e-mail: zmliu@sjtu.edu.cn).}}

\markboth{IEEE TRANSACTIONS ON SYSTEMS, MAN, AND CYBERNETICS,~Vol.~14, No.~8, November~2026}%
{Zhao \MakeLowercase{\textit{et al.}}: Dual Representation Enhanced Diffusion Policy for Flexible Job Shop Scheduling}


\maketitle

\begin{abstract}
The Flexible Job-Shop Scheduling Problem (FJSP) is a fundamental and challenging optimization problem in smart manufacturing. Although recent deep reinforcement learning (DRL) approaches can learn effective scheduling policies through interaction with the environment, their performance often degrades as problem scale and processing-time uncertainty increase. In this paper, we propose \textsc{DredRL}, a \underline{d}ual \underline{r}epresentation \underline{e}nhanced \underline{d}iffusion-policy \underline{r}einforcement \underline{l}earning scheduler. \textsc{DredRL} first constructs enhanced dual representations of operations and machines through self- and cross-attention mechanisms to capture both intra-entity and inter-entity dependencies. A diffusion model is further employed as the policy network, generating scheduling decisions through iterative denoising conditioned on the encoded state to enhance policy expressiveness and robustness. Extensive experiments on synthetic, public, large-scale, and stochastic FJSP benchmarks demonstrate that \textsc{DredRL} consistently outperforms state-of-the-art learning-based schedulers across most configurations. Compared with SOTA, \textsc{DredRL} reduces makespan by up to $5.44\%$ on synthetic instances. It also achieves the best learning-based performance on three of four public benchmarks and outperforms the stochasticity-aware algorithms on most SD3 configurations. These results demonstrate that combining enhanced dual representations with diffusion-based policies provides an effective and robust approach for scalable FJSP scheduling in complicated environment.
\end{abstract}


\section{Introduction}
\label{sec:introduction}
\IEEEPARstart{S}{mart} manufacturing is becoming increasingly critical to the sustainable and resilient development of a nation. Singapore's manufacturing 2030 aims to grow the sector’s value-add by 50\% by leveraging AI, robotics, and IIoT to build a sustainable, high-tech, and less labor-intensive ecosystem \cite{rie2030_factsheet}. In smart manufacturing, Flexible Job-Shop Scheduling Problem (FJSP) is a fundamental combinatorial optimization problem, where each operation can be processed by one of multiple eligible machines with different processing times. This flexibility substantially enlarges the decision space compared with classical job-shop scheduling, making NP-hard FJSP even more complicated and practically challenging in large-scale, dynamic production environments. In real-world workshops, schedulers must repeatedly decide which operation--machine pair to dispatch under complex constraints such as precedence relations, machine availability, and heterogeneous processing capabilities, while pursuing objectives such as minimizing makespan and improving throughput. Consequently, developing fast and high-quality scheduling approaches that generalize across instance sizes and remain effective under uncertainty is critical for modern intelligent manufacturing systems.

FJSP scheduling has been extensively studied, and existing methods can be broadly divided into two stages. The first stage is dominated by traditional approaches, including exact optimization and heuristic algorithms. Exact methods such as mixed-integer linear programming (MILP) \cite{ozguven2010mathematical} can produce optimal solutions for sufficiently small FJSP instances, while rule-based dispatching strategies such as FIFO \cite{ho2005evolving} and its variants incorporate human expertise into decision making. OR-Tools \cite{ortools} is also a strong benchmark in this stage, using combinatorial optimization to generate high-quality, near-optimal schedules. However, these methods become increasingly impractical in real-world settings, as computational cost grows exponentially with problem scale and online decision making requirements. Heuristic algorithms, such as genetic algorithms \cite{rooyani2019efficient}, are relatively flexible, but they rely heavily on manual design, parameter tuning, and domain knowledge, and their performance can also degrade as problem size increases. As a result, traditional methods face clear limitations on dynamic large-scale problems.

To overcome these issues, the second stage has shifted toward machine learning, particularly deep reinforcement learning (DRL). DRL is well suited to FJSP, since the problem can be formulated as a Markov decision process (MDP) in which selecting a feasible operation--machine pair corresponds to an action, while minimizing makespan can be directly incorporated into reward optimization. By learning scheduling policies through interaction with the environment, DRL enables fast inference after training and offers stronger adaptability to dynamic conditions and unseen instances \cite{ni2021multi}. Recent studies further combine DRL with graph neural networks (GNNs) and attention-based architectures to improve state representation. For example, HGNN \cite{song2022flexible} employs heterogeneous graph embeddings to capture structural relations among operations and machines, while DANIEL \cite{wang2023flexible} achieves state-of-the-art (SOTA) performance through attention-based state representation and PPO-Clip optimization \cite{huang2024ppo}. More recently, SPM-DAN extends learning-based FJSP scheduling to stochastic processing times by introducing a scenario processing module that explicitly encodes multiple processing-time scenarios and their associated uncertainty into the scheduling policy \cite{smit2025neural}. Such graph- and attention-based representations are particularly effective for FJSP because they capture structural dependencies and operation--machine interactions, thereby improving the state representation and decision quality of DRL agents. Related work has also explored multi-agent reinforcement learning (MARL) to enhance cooperation among agents \cite{zhang2023deepmag}.

While DRL models enable fast scheduling in FJSP, existing schedulers remain limited as states are often overly simplified and the scheduling complexity is not well captured. First, representing the intricate job shop structure remains difficult. The effectiveness of learning-based methods relies heavily on the quality and comprehensiveness of state representations \cite{xiao2025geld}. Operations and machines form a heterogeneous, bipartite dependency system in which both intra-entity dynamics and inter-entity couplings jointly decide the performance. Second, most DRL policies are parameterized by standard discriminative networks that output pointwise action distributions. Such policies cannot be sufficiently expressive for the multi-modal decision landscape of FJSP, particularly on larger instances. Third, existing DRL schedulers often struggle with model generalization and transferability. Many DRL methods are trained on a fixed distribution of instances and implicitly assume consistent state/action dimensions and similar feature statistics, which leads to overfitting to the training scale and layout. Even though several works have proposed complex state encoding schemes \cite{song2022flexible,lei2022multi,wang2023flexible}, the resulting representations can still be ambiguous in capturing complex scheduling semantics, and the learned policies often degrade when transferred to unseen tasks or instance distributions. Therefore, additional mechanisms are needed to better couple features across heterogeneous entities and to improve the model’s generalization capability.

In this paper, we propose \textsc{DredRL}, a novel \underline{d}ual \underline{r}epresentation \underline{e}nhanced \underline{d}iffusion policy \underline{r}einforcement \underline{l}earning scheduler for FJSP. \textsc{DredRL} combines a dual-representation encoder with a diffusion-based policy to improve scalability, expressiveness, and robustness. The encoder models operations and machines as two coupled streams, using enhanced self-attention to capture intra-entity dependencies and cross-attention to model operation--machine interactions. Based on the encoded state, the diffusion actor generates scheduling decisions through iterative denoising within an actor--critic DRL framework, enabling more expressive modeling of complex decision distributions. Extensive experiments on synthetic, public, large-scale, and stochastic benchmarks show that \textsc{DredRL} outperforms SOTA learning-based schedulers across most configurations. Compared with DANIEL, it reduces makespan by up to $5.44\%$ on small-scale and $3.53\%$ on larger-scale instances without size-specific retraining. It also achieves the best learning-based performance on most public benchmarks and SD3 stochastic settings, demonstrating strong cross-size, cross-distribution, and stochastic generalization.

The main contributions of this work are summarized as follows:

\begin{enumerate}
    \item We propose \textsc{DredRL}, a dual-representation-enhanced diffusion-policy DRL framework for FJSP that improves solution quality, scalability, and robustness across deterministic and stochastic scheduling environments.
    
    \item We design an enhanced dual representation encoder that jointly models operation and machine features, enabling richer inter-entity and intra-entity feature coupling for more effective operation--machine decision making.
    
    \item We develop a distinct diffusion-based actor formulation that employs a diffusion model as the policy network and generates scheduling decisions through iterative denoising conditioned on the encoded state, improving policy expressiveness and robustness under varying problem scales and processing-time uncertainty.
    
    \item We conduct extensive experiments covering synthetic, public, large-scale, and stochastic FJSP benchmarks. The results demonstrate consistent improvements over SOTA learning-based baselines, strong cross-size and cross-distribution generalization, and competitive performance under stochastic processing times, including improvements over the stochasticity-aware algorithms and reduced gaps to time-limited OR-Tools references.
\end{enumerate}


\section{Proposed Method}

\begin{figure*}[h!]
\includegraphics[width=\textwidth]{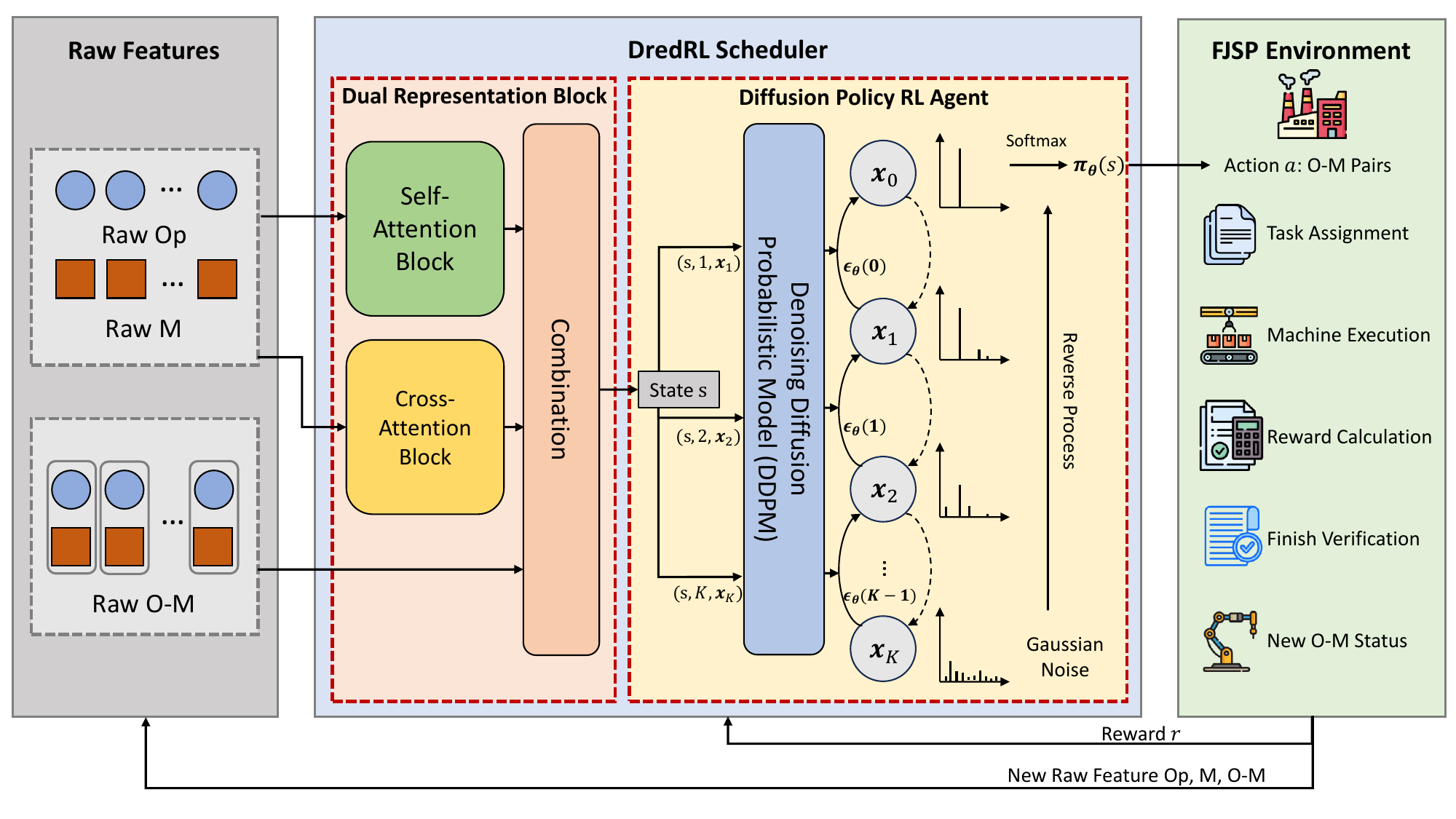}
\caption{Overall system architecture of the proposed \textsc{DredRL} scheduler for FJSP. Raw operation, machine, and operation--machine pairing features are encoded by a dual-representation module to form the state embedding $s$. A diffusion-policy RL agent then samples an action distribution $\pi_\theta(s)$ and selects compatible pairs for task assignment. The FJSP environment executes the assignment, computes reward $r$, updates status, and feeds back new raw features for the next decision step.}
\label{img1:sysarch}
\end{figure*}

FJSP is a classical optimization problem that is known to be NP-hard. The FJSP can be formulated into a decision-making process by adopting an MDP. \textsc{DredRL} combines a dual-representation encoder with a diffusion-based policy as shown in Figure~\ref{img1:sysarch}. In this section, we will introduce preliminaries of FJSP, our proposed method and system configurations including MDP formulation and a novel dual representation enhanced diffusion policy for FJSP. 

\subsection{Flexible Job-shop Scheduling Problem}

The FJSP usually considers a set of $n$ jobs and $m$ machines, which are represented by $\mathcal{J} = \left\{{J}_1, {J}_2 , \dots, {J}_n\right\}$ and $\mathcal{M} = \left\{{M}_1, {M}_2 , \dots, {M}_m\right\}$ respectively. Each job $\mathcal{J}_i$ consists of a sequence of $n_i$ operations noted by $\mathcal{O}_i = \left\{{O}_{i,1}, \dots, {O}_{i,n_i} \right\}$. With an assumption that all jobs have arrived at starting time $T=0$, each operation ${O}_{i,j}$ can be processed by a subset of feasible machines ${M}_{i,j} \subseteq \mathcal{M}$. The processing time of operation ${O}_{i,j}$ on machine ${M}_{k} \in {M}_{i,j}$ can be described as $p^{k}_{i,j} > 0$. The optimization of FJSP aims at operation-machine assignments while obeying the precedence constraints of operations and machine availability. The preliminary target of FJSP optimization is to minimize the completion time of all jobs (makespan).

A well-known method to represent FJSP is the disjunctive graph. It is considered a clearer way of formulating FJSP with both operation precedence and machine availability constraints. A disjunctive graph $\mathcal{G}$ is described by a tuple $(\mathcal{V}, \mathcal{C}, \mathcal{D})$. The node set $\mathcal{V} = \mathcal{O} \cup \left\{\textrm{Start}, \textrm{End}\right\}$ represents all operation nodes $\mathcal{O}$ and starting, terminating nodes. The conjunction set $\mathcal{C} = \left\{\langle{O}_{i,j}, {O}_{i,j+1}\rangle\right\}$ consists of directed edges between operation nodes of each job ${J}_i$, following the precedence constraints of operations. The disjunction set $\mathcal{D} = \left\{{D}_1, \dots, {D}_m\right\}$ comprises $m$ groups, reflecting that each machine can possibly process a subset of operations. A solution to an FJSP is described by a directed acyclic graph in which each operation ${O}_{i,j}$ has been scheduled to a specific machine ${M}_{k}$ with remaining disjunctions lined to ${O}_{i,j}$ deleted. 

\subsection{Dual Representation Network}

\begin{figure*}[h!]
\includegraphics[width=\textwidth]{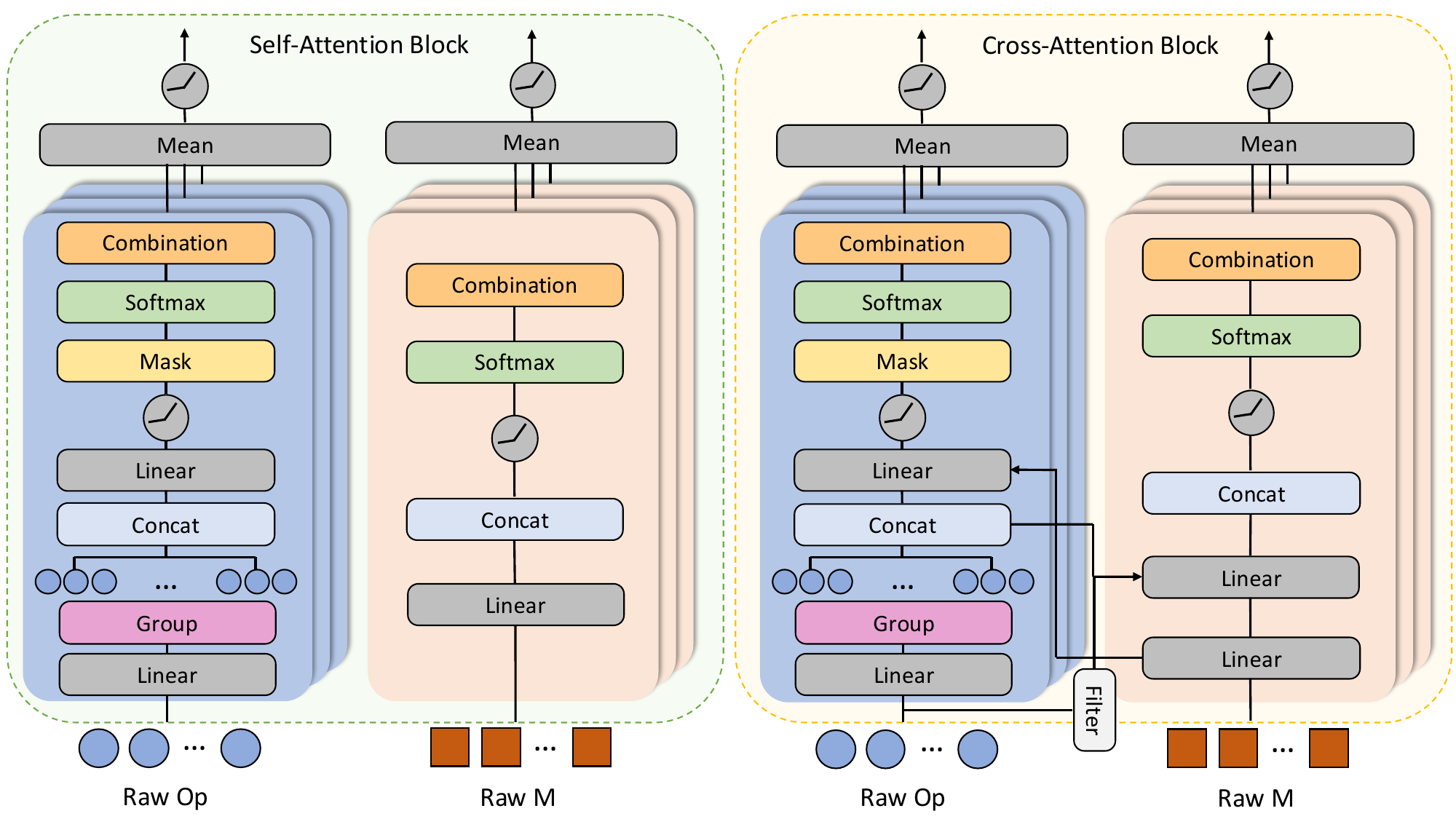}
\caption{Dual-representation encoder with self- and cross-attention blocks. Raw operation (Op) and machine (M) features are linearly projected and processed via masked graph attention, followed by mean pooling and ELU to derive embeddings.}
\label{img2:dualrep}
\end{figure*}

In complex environments like FJSP, multiple types of relationship exist among different operations and machines. Adopting self-attention models like Graph Attention Networks (GATs) \cite{velickovic2018graphattentionnetworks} in many existing solutions have been proved to conduce to feature extraction. However, relying entirely on one type of attention mechanism may create blind spots for the DRL agent, neglecting the inner connection between different features. To overcome this issue, we propose a novel dual representation model for \textsc{DredRL}, involving both intra- and inter-entity (dual) representations of complicated features in FJSP environment, as shown in Figure~\ref{img2:dualrep}. Details will be introduced below. GATs are one of the effective and commonly used GNN architectures with self-attention. A multi-head attention mechanism is applied to both attention blocks to enhance their learning ability \cite{vaswani2023attentionneed} and the multi-head attention parameter is denoted as $r\in\{1,\ldots,H\}$.

\subsubsection{Self-attention Block.} Self-attention models excel in relating input elements inside the same entity. This step helps the \textsc{DredRL} to learn the intra-entity representation. In some existing FJSP solutions, GATs are applied as self-attention module to update node features by attending over neighbors. In this intra-entity representation learning step, operation features and machine features are processed separately with local self-attention block. The operation feature block helps find out the significance of queuing operations, while the machine feature block leads to the competition between multiple machines based on their compatibility of operations. By stacking $L>1$ graph attention layers, the GAT yields final node embeddings for downstream steps.

\paragraph{Operation Nodes.}
Let $\mathcal{O}$ and $\mathcal{M}$ denote the operation and machine sets at step $t$. For both blocks, given input node features $\mathcal{H}=\{{\mathbf h}_i\in\mathbb{R}\}_{i=1}^{N}$, the GAT finally output a feature embedding $\tilde{ \mathbf h}^{\prime}_{i}$. For operation ${O}_{i,j}$, a neighbor set $\mathcal{N}_{{O}_{i,j}} = \left\{{{O}_{{i,j}}}, {{O}_{{i,j-1}}}, {{O}_{{i,j+1}}}\right\}$ includes neighbor nodes of operation ${O}_{i,j}$ (including itself). The attention coefficients between the neighbor nodes represent the relationships between operation ${O}_{i,j}$ itself, its predecessor ${O}_{i,j-1}$ and successor ${O}_{i,j+1}$ if they exist. Each GAL computes attention coefficients for operation node ${O}_{i,j}$ as,
\begin{equation}
\begin{aligned}
e^{\mathrm{op}}_{i,j_1,j_2}
&=\mathrm{LeakyReLU}\!\left(
\bar{\bm{a}}_{\mathrm{op}}^{\top}
\big[
(\mathbf {W}^{\mathrm{op}} \mathbf h_{{O}_{i,j_1}})\,\|\,(\mathbf {W}^{\mathrm{op}} \mathbf h_{{O}_{i,j_2}})
\big]
\right),\\
&\qquad j_1,j_2\in \{j-1,\,j,\,j+1\}.
\end{aligned}
\end{equation}
where ${\mathbf h}_{{O}_{i,j}}, {\mathbf h}_{{O}_{i,j-1}}, {\mathbf h}_{{O}_{i,j+1}}$ are input graph feature vectors after grouping, $\mathbf {W}^{\mathrm{op}}$ is a linear transformation matrix shared by all operation nodes, the operation $[\cdot\|\cdot]$ concatenates two vectors to a unified vector, and $\bar{\bm{a}}_{\mathrm{op}}$ is the weight vector of a single-layer network with a $\mathrm{LeakyReLU}(\cdot)$ activation. 
A dynamic mask on operation node attention coefficients is introduced to block invalid operations at each step. For each operation $O_{i,j}$, we define its attention neighborhood as the operation itself and its valid immediate predecessor and successor within the same job. We apply masking via,
\begin{equation}
\Gamma_{O_{i,j}, O_{i',j'}}(t)=
\begin{cases}
0, & O_{i',j'}\in \mathcal{N}_{O_{i,j}},\\
-\infty, & \textrm{otherwise}.
\end{cases}
\label{eq:mask}
\end{equation}

The attention coefficients are normalized via a softmax function as,
\begin{equation}
\alpha_{i,j,k}=\frac{\exp(e_{i,j,k})}{\sum_{k^{\prime}=j-1}^{k^{\prime}=j+1}\exp(e_{i,j,k^{\prime}})}.
\end{equation}

The weighted sum of transformed neighboring features is computed by,
\begin{equation}
\widetilde{\mathbf h}^{(\mathrm{op})'}_{\mathcal{O}_{i,j}}=\sigma\!\left(\sum_{k=j-1}^{j+1}\alpha_{i,j,k}\mathbf {W}^{\mathrm{op}}{\mathbf h}_{{O}_{i,k}}\right),
\end{equation}
where $\sigma$ is a nonlinear activation, typically $\mathrm{ELU}(\cdot)$ function. We combine multi-heads by concatenation:
\begin{equation}
\label{eq:multi-heads}
\widetilde{\mathbf h}_{{O}_{i,j}}^{\mathrm{op}}=
\mathrm{Concat}\!\left(\widetilde{\mathbf h}_{{O}_{i,j}}^{\mathrm{op},(1)},\ldots,\widetilde{\mathbf h}_{{O}_{i,j}}^{\mathrm{op},(H)}\right).
\end{equation}

\paragraph{Machine Nodes.} In machine self-attention block, the calculation is focused to extract machine feature relationship between machine nodes. For machine node $\mathcal{M}_i$, a neighbor set of $\mathcal{N}_{{M}_i} = \left\{j | \forall C_{i,j} \neq \emptyset \right\}$ denotes neighbor nodes, where $C_{i,j}$ represents an operation set competed by machine ${M}_i$ and ${M}_j$. In this way, self-attention mechanism in machine entity allows multiple machines competing over their commonly compatible operations. The attention coefficients between nodes ${M}_{i}$ and ${M}_{j}$ are given by,
\begin{equation}
e^{\mathrm{m}}_{i,j}=\mathrm{ LeakyReLU}\!\left(\bar{\bm{a}}_{\mathrm{m}}^{\top}\big[(\mathbf {W}^{\mathrm{m}}{\mathbf h}_{{M}_{i}})\,\|\,(\mathbf {W}^{\mathrm{m}}{\mathbf h}_{{M}_{j}})\big]\right),
\end{equation}
where $\mathbf {W}^{\mathrm{m}}$ is a linear transformation matrix between machine nodes, and $\bar{\bm{a}}_{\mathrm{m}}$ is the weight vector of a single-layer network for machine self-attention block. Similar softmax activation, combination steps are then introduced to derive output embeddings as,
\begin{equation}
\widetilde{\mathbf h}^{(\mathrm{m})'}_{{M}_{i}}=\sigma\!\left(\sum_{k\in\mathcal{N}_{\mathcal{M}_i}}\alpha_{i,k}\mathbf {W}^{\mathrm{m}}{\mathbf h}_{{M}_{k}}\right).
\end{equation}

\subsubsection{Cross-attention Block.} Cross-attention models are beneficial for correlating features from different entities, which conduce to the learning of inter-entity representation. We focus on capturing the operation-machine relationship by using cross-attention block. Although there are operation-machine pairing features to address the connection, each operation and its compatible machines, the inter-entity relationship between operation and machine features should be further represented in a more expressive way. Therefore, we developed cross-attention block to capture the in-depth connections between operations and machines. Specifically we use the following process for cross-attention calculation. 

\paragraph{Operation-Machine Block.} We denote node input features of operation ${O}_{i,j}$ as
$\{\mathbf h_{{O}_{i,j}}\}$ and machine ${M}_k$ as $\{\mathbf h_{{M}_k}\}$.
We denote $\mathcal{A}(t)\subseteq \mathcal{O}\times \mathcal{M}$ as the compatibility edge set, and $\mathcal{N}_{i,j} =\{k \mid ({O}_{i,j}, {M}_k)\in\mathcal{A}(t)\}$ the compatible machines for operation ${O}_{i,j}$. The multi-head attention parameter is $r\in\{1,\ldots,H\}$ with output dimension $d_h$. 
For each compatible pair $({O}_{i,j}, {M}_k)\in\mathcal{A}(t)$,
\begin{equation}
e_{i,j,k}^{M\rightarrow O}=\mathrm{LeakyReLU}\!\left(
\bm{a}_{M\rightarrow O}^\top \left[\mathbf {Z}^{\mathrm{op}} \mathbf h_{{O}_{i,j}} \,\|\, \mathbf {Z}^{\mathrm{m}}  \mathbf h_{{M}_k} \right]
\right),
\end{equation}
\noindent where $Z^{op}\in\mathbb R^{d_h\times d_o}$ and $Z^{m}\in\mathbb R^{d_h\times d_m}$ project the two entity types into compatible feature spaces, $||$ denotes vector concatenation, and $\bm{a}_{M\to O}\in\mathbb R^{2d_h}$ is learned. This is an additive graph-attention score: it incorporates both operation and machine information. The cross attention mask will be described as, 

\begin{equation}
 \Gamma_{i,j,k}^{OM}(t)
 =\begin{cases}
 0,&(O_{i,j},M_k)\in\mathcal{A}(t),\\
 -\infty,&\text{otherwise}.
 \end{cases}
 \label{eq:OM-mask}
\end{equation}

After applying mask, we normalize over the neighborhood $\mathcal{N}_{i,j}$ as,
\begin{equation}
 \alpha_{i,j,k}^{M\to O}
 =\frac{\exp(e_{i,j,k}^{M\to O})}
 {\displaystyle\sum_{k'\in\mathcal{N}_{i,j}^{M}}
  \exp(e_{i,j,k'}^{M\to O})}.
 \label{eq:orig-MO-alpha}
\end{equation}
The cross-attended representation of each compatible pair $(\mathcal{O}_{i,j}, \mathcal{M}_k)\in\mathcal{A}(t)$ for each head is,
\begin{equation}
 \widetilde{\mathbf h}_{O_{i,j}}^{M\to O}
 =\sigma\!\left(
 \sum_{k\in\mathcal{N}_{i,j}^{M}}
 \alpha_{i,j,k}^{M\to O}
 \mathbf W_m^V\mathbf h_{M_k}
 \right),
\end{equation}
\noindent $\mathbf W_m^V\in\mathbb R^{d_v\times d_m}$ projects machine information into the space used to update operations. The superscript $V$ denotes a value projection. Multi-heads are concatenated as Eq.~(\ref{eq:multi-heads}).

\paragraph{Machine-Operation Block.} Similarly as attention block, for machine node ${M}_k$, a neighbor set of $\mathcal{N}_{{M}_k} = \left\{q | \forall C_{k,q} \neq \emptyset \right\}$ denotes neighbor nodes, where $C_{k,q}$ represents an operation set competed by machine ${M}_k$ and ${M}_q$. We define $\mathcal{N}_{k} =\{(i,j) \mid ({O}_{i,j}, {M}_k)\in\mathcal{A}(t)\}$ to be the compatible operations for selected machine ${M}_{k}$ and $\mathcal Q_k$ to denote the machine neighbors of $M_k$ participating in this competition model. Let $J_c(t)$ denote the candidate operation set. With reference to \cite{wang2023flexible} we introduce $\mathbf c_{kq} = \sum_{O_{ij} \in C_{kq} \cap J_c} \mathbf h_{O_{ij}}$ to measure the competition intensity between $M_k$ and $M_q$. A filter is applied to extract $\mathbf c_{kq}$ from input operation features. For each compatible pair $({O}_{i,j}, {M}_k)\in\mathcal{A}(t)$ with neighbors $\mathcal{N}_{{M}_k}$, the cross-attention coefficients are calculated as,
\begin{equation}
\begin{aligned}
e_{i,j,k,q}^{O\rightarrow M}&=\mathrm{LeakyReLU}\!\bigg(
\bm{a}_{O\rightarrow M}^\top \\
& \qquad \left[\mathbf {Z}^{\mathrm{op}} \mathbf h_{{O}_{i,j}} \,\|\, \mathbf {Z}^{\mathrm{m}}  \mathbf h_{{M}_k} \,\|\, \mathbf Z^c \mathbf c_{k,q}\right]
\bigg),
\end{aligned}
\end{equation}

\noindent where $\mathbf Z^c$ stands for an additional weight matrix for $\mathbf c_{kq}$. The third projected feature describes competition between the receiving machine $M_k$ and another machine $M_q$. Because three projected $d_h$-dimensional vectors are concatenated, $\bm a_{O\to M}\in\mathbb R^{3d_h}$. After applying the cross attention mask as Eq. \eqref{eq:OM-mask}, the normalization will be, 

\begin{equation}
 \beta_{i,j,k,q}^{O\to M}
 =\frac{\exp(e_{i,j,k,q}^{O\to M})}
 {\displaystyle
  \sum_{(i',j')\in\mathcal{N}_k}\sum_{q'\in\mathcal{Q}_k}
  \exp(e_{i',j',k,q'}^{O\to M})}.
 \label{eq:competition-alpha}
\end{equation}

After softmax, another machine-operation cross-attended representation of each compatible pair $({O}_{i,j}, {M}_k)\in\mathcal{A}(t)$ is,
\begin{equation}
 \widetilde{\mathbf h}_{M_k}^{O\to M}
 =\sigma\!\left(
 \sum_{(i,j)\in\mathcal{N}_k}
 \sum_{q\in\mathcal{Q}_k}
 \beta_{i,j,k,q}^{O\to M}
 \mathbf W_{op}^V\mathbf h_{O_{i,j}}
 \right),
 \label{eq:competition-minimal}
\end{equation}

\noindent where $\mathbf W_{op}^V\in\mathbb R^{d_v\times d_o}$ projects operation features into the machine space.

\subsubsection{Combination of Dual Representation}

We denote that after $L$ GAT layers, the features are processed from $\mathbf h^{(0)}$ to $\mathbf h^{(L)}$. We first apply average pooling to operation and machine embeddings as

\begin{subequations}
\begin{align}
    & \mathbf h_{O_{i,j}}^{(L)} = \mathrm{Mean}\Big( \widetilde{\mathbf h}^{(\mathrm{op}, L)}_{{O}_{i,j}}, \widetilde{\mathbf h}^{({M\to O},L)}_{{O}_{i,j}}\Big), \\
    & \mathbf h_{M_k}^{(L)} = \mathrm{Mean}\Big( \widetilde{\mathbf h}^{(\mathrm{m}, L)}_{{M}_{k}}, \widetilde{\mathbf h}^{(O\to M,L)}_{{M}_{k}}\Big), \\
    & \mathbf h_G^{(L)} = \left[ \left( \frac{1}{|\mathcal{O}_u|} \sum_{O_{i,j} \in \mathcal{O}_u} \mathbf h_{O_{i,j}}^{(L)} \right) \;\Big\|\; \left( \frac{1}{|\mathcal{M}_u|} \sum_{M_k \in \mathcal{M}_u} \mathbf h_{M_k}^{(L)} \right) \right],
\end{align}
\end{subequations}

\noindent where $\mathbf h_{O_{i,j}}^{(L)}$ stands for combined operation embeddings at layer $L$, and $\mathbf h_{M_k}^{(L)}$ means combined machine embeddings at layer $L$, and lastly $\mathbf h_G^{(L)}$ represents concatenated global FJSP features at layer $L$.

\subsection{Deep Reinforcement Learning}

As our target is to dynamically schedule operations to compatible and available machines and update the disjunctive graph $\mathcal{G}$ step by step, an MDP is formulated to enable recursive interaction between scheduling agent and the FJSP environment. At each time step $t$, the scheduling agent receives a state $s_t$ from the FJSP environment and chooses an action $a_t$ to assign the operation to a compatible machine for processing. The FJSP environment processes the action $a_t$ and returns a reward $r_t$ accordingly. By repeating this process, the disjunctive graph will be updated step by step and eventually becomes the solution map. The MDP is defined as the follows.

\paragraph{State}
At step $t$, we only keep relevant operations and machines for decision-making: completed operations and machines that cannot process any remaining operation are ignored. Let $\mathcal{O}_u(t)$ and $\mathcal{M}_u(t)$ denote the relevant operation and machine sets, and $\mathcal{A}(t)$ denotes the compatible operation-machine pairs. The state is a set of feature vectors
$h_{O_{ij}}\!\in\!\mathbb{R}$ ($O_{ij}\!\in\!\mathcal{O}_u(t)$),
$h_{M_k}\!\in\!\mathbb{R}$ ($M_k\!\in\!\mathcal{M}_u(t)$),
and $h_{(O_{ij},M_k)}\!\in\!\mathbb{R}$ ($(O_{ij},M_k)\!\in\!\mathcal{A}(t)$).
As scheduling progresses, the state set shrinks accordingly. We algin with existing solutions to derive operation, machine, and operation-machine pair features. We follow the state construction of \cite{wang2023flexible}.

\paragraph{Action}
The action space is $\mathcal{A}(t)$, comprised of all compatible operation-machine pairs $(O_{i,j},M_k)$ defined at step $t$. All operations in $\mathcal{A}(t)$ are selected from the candidate set $\mathcal{J}_c(t)$.

\paragraph{State transition}
Executing $a_t$ starts processing an operation on a machine, updates the environment, and yields the next state $s_{t+1}$ by updating $\mathcal{O}_u(t)$, $\mathcal{M}_u(t)$, $\mathcal{A}(t)$ and computing features.

\paragraph{Reward}
We choose the negative value of the estimated makespan as a reward $r_t$ at each step $t$. Inspired by \cite{zhang2020learning, wang2023flexible}, at each step $t$, we similarly estimate the lower bound of the completion time for each operation $O_{i,j}$ as $\underline{C}(O_{i,j},s_t)$. If operation $O_{i,j}$ is scheduled, its value is equivalent to the actual completion time $C_{i,j}$. Otherwise, it is approximated by the sum of the lower bound of previous operation $\underline{C}(O_{i,j-1}, s_t)$ and the minimal processing time of $O_{i,j}$ among all feasible machines $M_{i,j}$ as, 
\begin{equation}
\underline{C}(O_{i,j}, s_t)=
\begin{cases}
 C_{i,j}(s_t),&\textrm{scheduled},\\
 \underline{C}(O_{i,j-1}, s_t)+\displaystyle\min_{k\in{M}_{i,j}} p^k_{ij},& \textrm{not scheduled}.
 \end{cases}
\end{equation}
We note that $\underline{C}(O_{i,0})=0$ and the maximum value of the completion time across all operations ${O_{i,j}\in\mathcal{O}}$ will be the makespan at step $t$. Therefore, we represent the step-wise reward as the difference of,
\begin{equation}
r_t=\max_{O_{i,j}\in\mathcal{O}} \underline{C}(O_{i,j}, s_t)-\max_{O_{i,j}\in\mathcal{O}} \underline{C}(O_{i,j}, s_{t+1}).
\end{equation}
With the discount factor $\gamma=1$, maximizing the return $\sum_t r_t$ is equivalent to minimizing the makespan $C_{\max}$.

\paragraph{Policy}
We apply an actor-critic framework, specifically with a diffusion model-based policy $\pi(a_t\mid s_t)$ over $a_t\in\mathcal{A}(t)$. The details will be introduced in the following section.

\subsection{Basics of Diffusion Models}

The actor network of \textsc{DredRL} is based on a novel design and built upon a diffusion model that generates actions by iteratively denoising Gaussian noise conditioned on the processed state embeddings, enabling expressive action distributions under uncertainty \cite{zhao2026diffusion}. Below we summarize the forward noising and reverse denoising processes in denoising diffusion probabilistic model (DDPM).

\subsubsection{Forward Process}
The forward diffusion process gradually adds Gaussian noise to $\bm{x}_0$, producing $\bm{x}_1,\ldots,\bm{x}_K$:
\begin{equation}
\label{eq3-5:diff-forward-q}
q(\bm{x}_k|\bm{x}_{k-1})=\mathcal{N}\!\left(\bm{x}_k\mid \sqrt{1-\beta_k}\bm{x}_{k-1},\,\beta_k\bm{I}\right),
\end{equation}
where $\beta_k$ is scheduled for $k=1,\ldots,K$. Following \cite{ho2020denoising}, we use the variational posterior scheduler,
\begin{equation}
\label{eq3-6:diff-forward-beta}
\beta_k = 1 - e^{-\frac{\beta_{\min}}{K} - \frac{2k - 1}{2K^2}(\beta_{\max} - \beta_{\min})},
\end{equation}
with hyper-parameters $\beta_{\min}$ and $\beta_{\max}$. The reverse conditional posterior (given $\bm{x}_0$) is,
\begin{equation}
\label{eq3-8:diff-forward-reverse-condition}
q(\bm{x}_{k-1}|\bm{x}_k, \bm{x}_0) = \mathcal{N}\!\left(\bm{x}_{k-1}; \tilde{\bm{\mu}}(\bm{x}_k, k), \tilde{\beta}_k \bm{I}\right).
\end{equation}
As given in \cite{ho2020denoising}, the forward process can be regarded as a Markov provess, the distribution $\bm{x}_K$ given $\bm{x}_0$ can be formulated the product of transitions $q(\bm{x}_k | \bm{x}_{k-1})$ over denoising steps as,
\begin{equation}
\label{eq3-10:diff-forward-transitions}
    q(\bm{x}_{1:K} | \bm{x}_0) = \prod_{k=1}^{K} q(\bm{x}_k | \bm{x}_{k-1}).
\end{equation}

The forward process provides mathematical relationships between $\bm{x}_k$ and $\bm{x}_0$ among denoising step $k$, with a standard normal noise $\bm{\epsilon} \sim \mathcal{N}(0,\bm{I})$ by,
\begin{equation}
\label{eq3-10.5:diff-forward-relationships}
    \bm{x}_k = \sqrt{\bar{\alpha}_k}\bm{x}_0 + \sqrt{1-\bar{\alpha}_k}\bm{\epsilon},
\end{equation}

\noindent where $\alpha_k=1-\beta_k$. As $k$ increases to $K$, $\bm{x}_K$ finally becomes pure noise following a normal distribution of $\mathcal{N}(0,\bm{I})$.


\subsubsection{Reverse Process}
The reverse process starts from pure noise $\bm{x}_K\sim\mathcal{N}(0,\bm{I})$ and iteratively denoises to obtain $\bm{x}_0$. Since $p(\bm{x}_{k-1}|\bm{x}_k)$ is intractable, we approximate it by,
\begin{equation}
\label{eq3-11:diff-reverse-model-ptheta}
p_{\bm{\theta}}(\bm{x}_{k-1}|\bm{x}_k)=
\mathcal{N}\!\left(\bm{x}_{k-1};\,\bm{\mu}_{\bm{\theta}}(\bm{x}_k,k,s),\,\Sigma_{\bm{\theta}}(\bm{x}_k,k,s)\right),
\end{equation}
where $s$ is the state embeddings, including operation, machine and operation-machine pairing features after dual representation. With the scheduler, we set $\Sigma_{\bm{\theta}}(\bm{x}_k,k,s)=\tilde{\beta}_k\bm{I}$. After applying Bayesian rules, the reverse process could be reformatted into a Gaussian probability density function where the mean is described as,
\begin{equation}
\label{eq3-12: diff-reverse-gaussian-mean}
    \tilde{\bm{\mu}}(\bm{x}_k,k,s)=
\frac{\sqrt{\alpha_k}\left(1-\bar{\alpha}_{k-1}\right)}{1-\bar{\alpha}_k}
\bm{x}_k+\frac{\sqrt{\bar{\alpha}_{k-1}}\beta_{k}}{1-\bar{\alpha}_k}\bm{x}_0,
\end{equation}

\noindent where $k = 1, \dots, K$. We parameterize denoising via a noise-prediction network $\bm{\epsilon}_{\bm{\theta}}(\bm{x}_k,k,s)$ and reconstruct,
\begin{equation}
\label{eq3-13:diff-reverse-model-x0}
\hat{\bm{x}}_0=
\frac{1}{\sqrt{\bar{\alpha}_k}}\bm{x}_k
-\sqrt{\frac{1}{\bar{\alpha}_k}-1}\cdot
\tanh\!\left(\bm{\epsilon}_{\bm{\theta}}(\bm{x}_k,k,s)\right),
\end{equation}
which yields the corresponding mean estimate after applying to Eq. \eqref{eq3-12: diff-reverse-gaussian-mean} as,
\begin{equation}
\label{eq3-14:diff-reverse-model-mu-new}
\tilde{\bm{\mu}}(\bm{x}_k,k,s)=
\frac{1}{\sqrt{\alpha_k}}
\left(\bm{x}_k-\frac{\beta_k\,\tanh\!\big(\bm{\epsilon}_{\bm{\theta}}(\bm{x}_k,k,s)\big)}{\sqrt{1-\bar{\alpha}_k}}\right).
\end{equation}
Overall, the generation distribution is,
\begin{equation}
\label{eq3-15:diff-reverse-model-ptheta-x0}
p_{\bm{\theta}}(\bm{x}_0)=p(\bm{x}_K)\prod_{k=1}^{K} p_{\bm{\theta}}(\bm{x}_{k-1}|\bm{x}_k),
\end{equation}
with $p(\bm{x}_K)=\mathcal{N}(0,\bm{I})$.

\subsection{Decision Making Process}

We hereby design the decision making process based on a actor-critic framework, where two Multi-Layer Perceptrons (MLPs) parameterized by $\theta$ and $\phi$ are used as actor and critic respectively. Specifically, $\mathrm{MLP}_{\theta}$ is designed for DDPM. We denote a state embeddings $s$ after $L$ GAT layers as,
\begin{equation}\label{eq:state_def}
    s = \mathbf h_{O_{i,j}}^{(L)} \,\|\, \mathbf h_{M_k}^{(L)} \,\|\, \mathbf h_{G}^{(L)} \,\|\, \mathbf h_{({O}_{i,j}, {M}_{k})},
\end{equation}

\noindent where $\mathbf h_{({O}_{i,j}, {M}_{k})}$ is the operation-machine pairing features. The denoising process is repeatedly calculated through Eq.~(\ref{eq3-14:diff-reverse-model-mu-new}) and we may derive a scalar,
\begin{equation}
    \mu(a_t\mid s_t) = \mathrm{DDPM}_{\theta}(s_t).
\end{equation}

\noindent After applying a softmax function, a policy is further derived as,
\begin{equation}
    \pi_\theta (a_t \mid s_t) = \frac{\exp(\mu(a_t \mid s_t))}{\sum_{b_t \in \mathcal{A}(t)} \exp(\mu(b_t \mid s_t))}.
\end{equation}

The critic network $\mathrm{MLP}_{\phi}$ is an estimator which takes the global features $\mathbf h_G^{(L)}$ as input to produce a scalar $v_\phi(s_t)$ as an approximation of the state value. 
%
%

\subsection{Training}

We use an encoder $f_{\bm{\psi}}$, a diffusion actor $\pi_{\bm{\theta}}$ and a critic model $V_{\bm{\phi}}$, initialized by parameters $\Theta=\{\bm{\psi},\bm{\theta},\bm{\phi}\}$. The training target is to minimize the loss function regarding the parameters $\Theta$. After computing advantages $\hat{A}_t$, returns $\hat{R}_t$ by GAE$(\gamma,\lambda)$, and $\log\pi_{\bm{\theta}}(a_t\mid s_t)$ is thereby calculated using the diffusion actor with mask $\Gamma(t)$. And the probability ratio between old and updated policies will be denoted as,

\begin{equation}
    r_t(\bm{\theta}) = \exp\!\left(\log\pi_{\bm{\theta}}(a_t\mid s_t)-\log\pi_{\bm{\theta}_{\mathrm{old}}}(a_t\mid s_t)\right).
\end{equation}

A loss function in PPO-clip format is applied to train the proposed \textsc{DredRL}, with following components, 

\begin{subequations}
\label{eq:loss_in_parts}
\begin{align}
    &\mathcal{L}^{\mathrm{clip}}_t(\bm{\theta}) = \min\Big(r_t(\bm{\theta})\hat{A}_t,\;\mathrm{clip}(r_t(\bm{\theta}),1-\epsilon,1+\epsilon)\hat{A}_t\Big),\\
    &\mathcal{L}^{v}_t(\bm{\phi}) = \big(V_{\bm{\phi}}(s_t)-\hat{R}_t\big)^2,\\
    &\mathcal{L}^{\mathrm{ent}}_t(\bm{\theta}) = H\big(\pi_{\bm{\theta}}(\cdot\mid s_t)\big),
\end{align}
\end{subequations}




\noindent where $\epsilon$ is a PPO-clip parameter. And the combined loss function will be described as $\mathcal{L}_t({\Theta})$, 

\begin{equation}\label{eq:loss_func}
    \mathcal{L}_t(\Theta)=\mathcal{L}^{\mathrm{clip}}_t(\bm{\theta})-c_v\mathcal{L}^{v}_t(\bm{\phi})+c_{\mathrm{ent}}\mathcal{L}^{\mathrm{ent}}_t(\bm{\theta}),
\end{equation}

\noindent where $c_\mathrm{v}, c_\mathrm{ent}$ are value coefficient and entropy coefficient. Overall, the algorithm of \textsc{DredRL} can be described as Algorithm~\ref{algo:dredrl}.

\paragraph{Time Complexity} Let $|\mathcal{O}_u(t)|$ and $|\mathcal{M}_u(t)|$ denote the numbers of relevant operations and machines at decision step $t$, and let $|\mathcal{A}(t)|$ be the number of compatible operation--machine pairs. Since our dual-representation encoder explicitly maintains two smaller graphs, one over operations and one over machines, and couples them only through the compatibility edges, the total number of processed connections is calculated as $|\mathcal{O}_u(t)| + |\mathcal{M}_u(t)|^2$ and $|\mathcal{A}(t)|$. For the diffusion actor, one action decision requires $K$ reverse denoising steps. Let $C_{\theta}$ denotes the computational cost of one noise-prediction evaluation $\bm{\epsilon}_{\bm{\theta}}(\bm{x}_k,k,s_t)$ and $C_V$ describes the cost of critics, with $L$ attention layers and hidden size $d$, the per-step rollout complexity is, 
\begin{equation}
\begin{aligned}
    T_{\mathrm{step}}(t)=O\!\Big( & Ld\big(|\mathcal{O}_u(t)|+|\mathcal{M}_u(t)|^2+|\mathcal{A}(t)|\big)\\
    & +(KC_{\theta}+1)|\mathcal{A}(t)|+C_V\Big).
\end{aligned}
\end{equation}

Thus, the overall training time complexity will be calculated as, 
\begin{equation}
    T_{\mathrm{train}}=O\!\Big(E(1+K)\sum_{t=0}^{T-1}T_{\mathrm{step}}(t)\Big),
\end{equation}

\noindent where $E$ is the total episode, and $T$ is the trajactory length.

\paragraph{Space Complexity} The algorithm stores one on-policy trajectory of length $T$, containing $(s_t,a_t,r_t,v_t,\log\pi_{\bm{\theta}_{\mathrm{old}}}(a_t\mid s_t),\Gamma(t))_{t=0}^{T-1}$. Let $d_s$ denote the dimension of $s_t$. Storing state embeddings costs $O(Td_s)$. If the candidate mask $\Gamma(t)$ is stored explicitly over $\mathcal{A}(t)$, it contributes $O\big(\sum_{t=0}^{T-1}|\mathcal{A}(t)|\big)$. Hence, the trajectory storage cost is, 
\begin{equation}
    M_{\mathrm{traj}}=O\!\Big(Td_s+\sum_{t=0}^{T-1}|\mathcal{A}(t)|\Big).
\end{equation}
In addition, model parameters require $O(|\bm{\psi}|+|\bm{\theta}|+|\bm{\phi}|)$ memory. Overall, the space complexity is, 
\begin{equation}
    M_{\mathrm{total}}=O\!\Big(Td_s+\sum_{t=0}^{T-1}|\mathcal{A}(t)|\Big)+O(|\bm{\psi}|+|\bm{\theta}|+|\bm{\phi}|).
\end{equation}

\begin{algorithm}[h]
\small
\caption{Training \textsc{DredRL} with PPO-Clip}
\label{algo:dredrl}
\KwIn{Environment $\mathcal{E}$; dual-representation encoder $f_{\bm{\psi}}$ with
cross-attention; per-candidate diffusion actor $\pi_{\bm{\theta}}$ with $K$ reverse
steps and prior scale $s_0$; critic $V_{\bm{\phi}}$; episodes $E$, horizon $T$,
clip $\epsilon$, PPO epochs $K_{\mathrm{ep}}$, GAT layer $L$, GAE $(\gamma,\lambda)$.}
\KwOut{$\bm{\psi}^*,\bm{\theta}^*,\bm{\phi}^*$}

Initialize $\Theta=\{\bm{\psi},\bm{\theta},\bm{\phi}\}$;
$\Theta_{\mathrm{old}}\!\leftarrow\!\Theta$\;
\For{$\mathrm{ep}=1$ \KwTo $E$}{
  \lIf{$\mathrm{ep}\bmod E_{\mathrm{rst}}=0$}{draw a fresh batch of training instances}
  Reset $\mathcal{E}$; observe raw features $\mathcal{H}_0$\;
  \For{$t=0$ \KwTo $T-1$}{
    Form candidate set $\mathcal{A}_t$ and mask $\Gamma_t$ from $\mathcal{H}_t$\;
    $(\mathbf h_{O_{i,j}}^{(L)} , \mathbf h_{M_k}^{(L)} , \mathbf h_{G}^{(L)})
      \leftarrow f_{\bm{\psi}_{\mathrm{old}}}(\mathcal{H}_t,\mathcal{A}_t)$\;
    Build per-candidate states $\{s_{t,c}\}_{c\in\mathcal{A}_t}$ by \eqref{eq:state_def}\;
    \For{$c\in\mathcal{A}_t$ \textnormal{(in parallel)}}{
      $\bm{x}^c_K\!\sim\!\mathcal{N}(\bm{0},s_0^{2}\bm{I})$\;
      $\bm{x}^c_0\leftarrow\mathrm{Reverse}^{\mathrm{det}}_{\bm{\theta}_{\mathrm{old}}}
        (\bm{x}^c_K,s_{t,c})$ by \eqref{eq3-13:diff-reverse-model-x0}\;
    }
    $\pi_{\bm{\theta}_{\mathrm{old}}}(\cdot\mid s_t)\leftarrow
      \mathrm{softmax}\big(g_{\bm{\theta}_{\mathrm{old}}}(\bm{x}^c_0)+\Gamma_t\big)$\;
    Sample $a_t\sim\pi_{\bm{\theta}_{\mathrm{old}}}(\cdot\mid s_t)$; store
      $(s_t,a_t,r_t,V_{\bm{\phi}_{\mathrm{old}}}(s_t),
       \log\pi_{\bm{\theta}_{\mathrm{old}}},\Gamma_t)$\;
    Step $\mathcal{E}$ to obtain $r_t,\mathcal{H}_{t+1}$\;
  }
  Compute $\hat{A}_t,\hat{R}_t$ by GAE$(\gamma,\lambda)$\;
  \For{$u=1$ \KwTo $K_{\mathrm{ep}}$}{
    \For{minibatch $\mathcal{B}$}{
      $r_t(\bm{\theta})\leftarrow\exp\!\big(\log\pi_{\bm{\theta}}(a_t\mid s_t)
        -\log\pi_{\bm{\theta}_{\mathrm{old}}}\big)$\;
      Update $\Theta$ by ascending $\mathcal{L}(\Theta)$ in \eqref{eq:loss_func}\;
    }
  }
  $\Theta_{\mathrm{old}}\!\leftarrow\!\Theta$\;
  \lIf{$\mathrm{ep}\bmod E_{\mathrm{val}}=0$}{validate on $\mathcal{E}_{\mathrm{val}}$}
}
\Return $\bm{\psi}^*,\bm{\theta}^*,\bm{\phi}^*$
\end{algorithm}

\vspace{-8pt}

\section{Results}

\subsection{Experimental Setup}

We present the experimental setup in the following aspects. Our experiments were conducted by a machine with an NVIDIA GeForce RTX 4090 GPU with 24GB memory and an Intel Core Ultra 7 265KF 20-Core CPU and 64GB of DDR5 RAM. The operating system is Ubuntu 24.04.2 LTS with CUDA V12.2 toolkit. Besides, our experiments are organized with the following details.


\subsubsection{Datasets}
For each FJSP configuration with $n$ jobs and $m$ machines, we generate 100 random instances for evaluation. Following existing works, we consider three types of synthetic datasets. The first dataset, denoted as SD1, follows \cite{song2022flexible}, where each job contains a varying number of sequential operations. For each operation $\mathcal{O}_{i,j}$ in an $n\times m$ instance, the number of compatible machines $|\mathcal{M}_{i,j}|$ and the corresponding processing time $p_{i,j}^{k}$ on machine $\mathcal{M}_k$ are sampled from $\mathcal{U}(1,m)$ and $\mathcal{U}(1,20)$, respectively. SD2 follows the same generation procedure but increases the processing-time range to $\mathcal{U}(1,99)$. To further evaluate robustness under processing-time uncertainty, we additionally adopt SD3 following SPM-DAN~\cite{smit2025neural}. SD3 retains the instance structure of SD2 while introducing correlated processing times across compatible machines. Specifically, a base processing time $p_{i,j}\sim\mathcal{U}(1,99)$ is first generated for each operation, after which the machine-specific processing time is sampled as $p_{i,j}^{k}\sim\mathcal{U}(0.85p_{i,j},1.15p_{i,j})$. To construct stochastic instances, each deterministic processing time is treated as the median, and a coefficient of variation $CV_{i,j}^{k}\sim\mathcal{U}(0.1,0.5)$ is sampled for each operation--machine pair, with $\sigma_{i,j}^{k}=CV_{i,j}^{k}p_{i,j}^{k}$. Following~\cite{smit2025neural}, log-normal processing-time distributions are used to represent stochastic realizations. For comprehensive evaluation, the deterministic SD1 and SD2 datasets are divided into two groups. The general-size group contains $10\times5$, $20\times5$, $15\times10$, and $20\times10$ instances, while the large-size group contains $30\times10$, $40\times10$, $100\times10$, and $200\times5$ instances. Models are trained and tested on the general-size datasets, whereas the large-size instances are used only for testing to evaluate cross-size generalization without size-specific retraining. For stochastic evaluation on SD3, we consider $10\times5$, $20\times5$, $15\times10$, and $20\times10$ instances and compare \textsc{DredRL} with deterministic and stochastic scheduling baselines under the same uncertainty setting.

\subsubsection{Hyperparameters} The actor and critic networks are optimized using learning rates $\alpha=3\times10^{-4}$. We set the discount factor to $\gamma=1$ to optimize the undiscounted episodic objective to ensure optimization of total makespan. The graph-attention encoder employs $L=2$ graph attention layers (GALs) with $H=4$ attention heads per layer. For the diffusion-based policy network, the DDPM denoiser runs for $K=10$ denoising steps. Training is conducted for at most $E=1000$ episodes, and the policy is validated every $E_{\mathrm{val}}=10$ episodes. The PPO-style loss uses coefficients $c_{\mathrm{p}}=1$ for the policy term and $c_{\mathrm{v}}=0.5$ for the value term, and an entropy regularization coefficient $c_{\mathrm{ent}}=0.01$ is applied to encourage exploration. Generalized advantage estimation (GAE) parameters $\lambda=0.98$ for SD1 and SD2 are applied to stabilize the training process. Finally, the hidden layers in actor and critic MLPs are parameterized as $\mathrm{MLP}_a=[64,64]$ and $\mathrm{MLP}_c=[64,64]$ respectively, representing two hidden layers with the specified numbers of neurons in each MLP. The PPO update epoch per batch is $K_{\mathrm{ep}}=4$ and minibatch size is $\mathcal{B}=1024$. For the stochastic SD3 experiments, additional parameters are introduced to characterize processing-time uncertainty. The base processing time is sampled from $[p_{\min},p_{\max}]=[1,99]$, and the processing time on each compatible machine is generated by applying a multiplicative perturbation $\delta\sim\mathcal{U}(0.85,1.15)$ to the corresponding base processing time. The coefficient of variation for each operation--machine pair is sampled as $CV\sim\mathcal{U}(0.1,0.5)$ to represent heterogeneous processing-time uncertainty. For stochastic decision evaluation, the Value-at-Risk confidence level is set to $\alpha_{\mathrm{VaR}}=0.95$. Unless otherwise stated, the remaining network architecture and DRL training settings are kept consistent with those used for SD1 and SD2. Details are further included in Table~\ref{tab3:training_param}.

\begin{table}[h!]
\centering
\caption{The configurations of \textsc{DredRL}.}
\label{tab3:training_param}
\renewcommand{\arraystretch}{1.3}
\begin{tabular}{ccc}
\hline\hline
\textbf{Parameters}       & \textbf{Description}                      & \textbf{Values}    \\ 
\midrule
\multicolumn{3}{l}{\textit{Architecture and training}}\\
$\alpha$                  & Learning rate                             & $3 \times 10^{-4}$ \\
$\gamma$                  & Discount factor                           & $1$                \\
$L$                       & GAL number                                & $2$                \\
$H$                       & Attention heads                           & $4$                \\
$K$                       & Denoising steps in DDPM                   & $10$               \\
$E$                       & Maximum training episodes                 & $1000$             \\
$\epsilon$                & PPO-clip parameter                        & $0.2$              \\
$c_{\mathrm{p}}$          & Policy coefficient                        & $1$                \\
$c_{\mathrm{v}}$          & Value coefficient                         & $0.5$              \\
$c_{\mathrm{ent}}$        & Entropy regularization coefficient        & $0.01$             \\
$\lambda$                 & GAE parameter                             & $0.98$             \\
$E_{\mathrm{val}}$        & Episodes between policy validations       & $10$               \\
$\mathrm{MLP}_a$          & Hidden layers and neurons of actor        & $[64,64]$          \\
$\mathrm{MLP}_c$          & Hidden layers and neurons of critics      & $[64,64]$          \\
$K_{\mathrm{ep}}$         & PPO update epochs per batch               & $4$                \\
$\mathcal{B}$             & Minimatch size                            & $1024$             \\
\midrule
\multicolumn{3}{l}{\textit{SD3 instance generation}}\\
$[p_{\min},p_{\max}]$      & Base processing-time range              & $[1,99]$ \\
$\delta$                   & Per-machine jitter on the base time     & $\mathcal{U}(0.85,1.15)$ \\
$|{M}_{ij}|$       & Compatible machines per operation       & $\mathcal{U}(1,5)$ \\
$CV$                       & Coefficient of variation per $(O_{ij},M_k)$ & $\mathcal{U}(0.1,0.5)$ \\
$\mathrm{VaR}_{\alpha}$    & Value at Risk (VaR) level                              & $0.95$ \\
\hline\hline
\end{tabular}
\end{table}

\subsubsection{Comparison Baselines} We evaluate \textsc{DredRL} against representative SOTA baselines and an optimization-based reference solver, OR-Tools, which can produce optimal or near-optimal schedules via combinatorial optimization. However, since FJSP is NP-hard, OR-Tools may require prohibitive runtime to certify optimality, particularly for large-scale instances. Following common practice, we therefore treat OR-Tools solutions as strong references rather than true optima, and impose a 30-minute time limit. We further compare with three recent learning-based SOTA methods, HGNN \cite{song2022flexible}, DANIEL \cite{wang2023flexible} and SPM-DAN \cite{smit2025neural}. HGNN formulates FJSP as a heterogeneous graph and leverages GNN message passing to capture relationships among scheduling entities, while DANIEL employs attention-based GAT for decision making. SPM-DAN further extends DANIEL to stochastic FJSP by introducing an attention-based scenario processing module that explicitly encodes sampled processing-time scenarios, enabling the policy to account for uncertainty during scheduling. For fairness, we report results from the original papers whenever available under their recommended settings; otherwise, we re-implement the methods to obtain missing results. Each result has been averaged from 10 independent runs due to the stochastic nature of algorithms.

\vspace{-8pt}

\subsection{Deterministic Results}

\subsubsection{Small and Large Size Instances} We first evaluate \textsc{DredRL} on small FJSP instances from SD1 and SD2 under both greedy and sampling decoding, as reported in Table~\ref{tab2:num_results}. Overall, \textsc{DredRL} consistently outperforms HGNN and remains competitive with or improves upon the SOTA learning-based baseline DANIEL. Considering the better decoding result of each method, \textsc{DredRL} achieves a lower makespan than DANIEL in 5 out of the 8 small-scale configurations, with an average makespan reduction of $1.46\%$ across these settings. The improvement is more pronounced on SD2, where \textsc{DredRL} achieves an average reduction of $2.66\%$ over DANIEL, compared with $0.25\%$ on SD1. This trend indicates a stronger advantage under the higher processing-time variability of SD2, where scheduling decisions become more challenging. In particular, on SD2 $15{\times}10$, \textsc{DredRL} reduces the best DANIEL makespan from $521.83$ to $493.42$, corresponding to a $5.44\%$ reduction, while decreasing the relative gap to OR-Tools from $38.35\%$ to $30.82\%$. A similar improvement is observed on SD2 $20{\times}10$, where the makespan decreases from $552.64$ to $531.39$ and the corresponding gap is reduced from $19.06\%$ to $14.48\%$. Although DANIEL remains slightly better on SD1 $20{\times}5$, SD1 $20{\times}10$, and SD2 $20{\times}5$, the differences are small, whereas \textsc{DredRL} provides more substantial gains on the more difficult SD2 cases. In addition, sampling generally improves the quality of the learned schedules: sampling produces a lower makespan than greedy decoding for \textsc{DredRL} in 14 of the 16 configurations in Table~\ref{tab2:num_results}. Several cases also yield negative gaps relative to OR-Tools, indicating that the learned policies can occasionally discover schedules with smaller makespan than those returned by the time-limited solver. Since OR-Tools is executed under a fixed computational budget of up to 1800 seconds and may terminate before reaching proven optimality, these negative gaps indicate that the learned policies can discover higher-quality schedules within only a few seconds of inference. This further indicates their computational efficiency for time-sensitive scheduling applications.

\begin{table}[t]
\centering
\caption{Performance comparison of different algorithms on synthetic datasets SD1 and SD2 with different instance sizes. Results are included with makespan and corresponding relative gap (\%) to OT-Tools.}
\label{tab2:num_results}
\renewcommand{\arraystretch}{1.3}
\resizebox{\linewidth}{!}{
\begin{tabular}{ccc|cc|cc|cc|c}
\hline\hline
\multicolumn{3}{c|}{\multirow{2}{*}{Size}}                         & \multicolumn{2}{c|}{HGNN\cite{song2022flexible}} & \multicolumn{2}{c|}{DANIEL\cite{wang2023flexible}} & \multicolumn{2}{c|}{\textsc{DredRL} (Ours)} & \multirow{2}{*}{OR-Tools} \\
\multicolumn{3}{c|}{}                                              & Greedy      & Sampling   & Greedy      & Sampling    & Greedy  & Sampling                &                           \\ \hline
\multirow{16}{*}{SD1} & \multirow{2}{*}{$10 \times 5$}  & Makespan & 111.67      & 105.59     & 106.71    & 101.67        & 106.24  & {\ul 99.79}             & 96.32                     \\
                      &                                 & Gap      & 15.94\%     & 9.62\%     & 10.79\%   & 5.55\%        & 10.30\% & {\ul 3.60\%}            & -                         \\ \cline{2-10} 
                      & \multirow{2}{*}{$20 \times 5$}  & Makespan & 211.22      & 207.53     & 197.56    & {\ul 192.78}  & 197.86  & 193.45                  & 188.15                    \\
                      &                                 & Gap      & 12.26\%     & 10.3\%     & 5.01\%    & {\ul 2.46\%}  & 5.16\%  & 2.82\%                  & -                         \\ \cline{2-10} 
                      & \multirow{2}{*}{$15 \times 10$} & Makespan & 166.92      & 161.28     & 160.86    & 153.22        & 160.54  & {\ul 152.86}            & 143.53                    \\
                      &                                 & Gap      & 16.30\%     & 12.37\%    & 12.07\%   & 6.75\%        & 11.85\% & {\ul 6.50\%}            & -                         \\ \cline{2-10} 
                      & \multirow{2}{*}{$20 \times 10$} & Makespan & 215.78      & 214.81     & 198.50    & {\ul 193.91}  & 199.82 & 195.30                  & 195.95                    \\
                      &                                 & Gap      & 10.10\%     & 9.61\%     & 1.29\%    & {\ul -1.06\%} & 1.97\% & -0.35\%                 & -                         \\ \cline{2-10} 
                      & \multirow{2}{*}{$30 \times 10$} & Makespan & 314.71      & 308.55     & 288.61    & 286.77        & 292.07 & {\ul 285.07}            & 274.67                    \\
                      &                                 & Gap      & 14.58\%     & 12.33\%    & 5.08\%    & 4.41\%        & 6.33\% & {\ul 3.79\%}            & -                         \\ \cline{2-10} 
                      & \multirow{2}{*}{$40 \times 10$} & Makespan & 417.87      & 410.76     & 379.28    & 379.71        & 384.20 & {\ul 378.14}            & 365.96                    \\
                      &                                 & Gap      & 14.18\%     & 12.24\%    & 3.64\%    & 3.76\%        & 4.98\% & {\ul 3.33\%}            & -                         \\ \cline{2-10}
                      & \multirow{2}{*}{$100 \times 10$}& Makespan & 1035.0      & 1023.6     & 946.0     & 985.3         & 930.5  & {\ul 918.1}              & 944.2                      \\
                      &                                 & Gap      & 9.62\%      & 8.41\%     & 0.19\%    & 4.35\%        & -1.45\%& {\ul -2.76\%}            & -                         \\ \cline{2-10} 
                      & \multirow{2}{*}{$200 \times 5$} & Makespan & 2055.1      & 2042.0    & 1901.1    & 2018.2        & 1903.05 & {\ul 1833.9}             & 1884.7                    \\
                      &                                 & Gap      & 9.04\%      & 8.35\%    & 0.87\%    & 7.08\%        & 0.97\% & {\ul -2.69\%}            & -                         \\ 
                      \hline
\multirow{16}{*}{SD2} & \multirow{2}{*}{$10 \times 5$}  & Makespan & 553.61      & 483.9      & 408.4     & 366.74         & 411.11  & {\ul 361.26}            & 326.24                    \\
                      &                                 & Gap      & 69.69\%     & 48.33\%    & 25.18\%   & 12.41\%        & 26.01\% & {\ul 10.73\%}           & -                         \\ \cline{2-10} 
                      & \multirow{2}{*}{$20 \times 5$}  & Makespan & 1059.04     & 962.9      & 671.03    & {\ul 629.94}   & 666.82  & 630.88                  & 602.04                    \\
                      &                                 & Gap      & 75.91\%     & 59.94\%    & 11.46\%   & {\ul 4.63\%}   & 10.76\% & 4.79\%                  & -                         \\ \cline{2-10} 
                      & \multirow{2}{*}{$15 \times 10$} & Makespan & 807.47      & 756.07     & 591.21    & 521.83         & 580.55  & {\ul 493.42}            & 377.17                    \\
                      &                                 & Gap      & 114.09\%    & 100.46\%   & 56.75\%   & 38.35\%        & 53.92\% & {\ul 30.82\%}           & -                         \\ \cline{2-10} 
                      & \multirow{2}{*}{$20 \times 10$} & Makespan & 1045.82     & 990.37     & 610.16    & 552.64         & 602.43  & {\ul 531.39}            & 464.16                    \\
                      &                                 & Gap      & 126.12\%    & 113.37\%   & 31.45\%   & 19.06\%        & 29.79\% & {\ul 14.48\%}           & -                         \\  \cline{2-10} 
                      & \multirow{2}{*}{$30 \times 10$} & Makespan & 1564.57     & 1486.56    & 794.62    & 757.48         & 785.19  & {\ul 749.53}            & 692.26                    \\
                      &                                 & Gap      & 126.01\%    & 114.74\%   & 14.79\%   & 9.42\%         & 13.42\% & {\ul 8.27\%}            & -                         \\ \cline{2-10} 
                      & \multirow{2}{*}{$40 \times 10$} & Makespan & 2048.96     & 1976.25    & 983.37    & 951.21         & 973.98  & {\ul 943.43}             & 998.39                    \\
                      &                                 & Gap      & 105.23\%    & 97.94\%    & -1.50\%   & -4.73\%        & -2.44\% & {\ul -5.50\%}           & -                         \\ \cline{2-10}
                      & \multirow{2}{*}{$100 \times 10$}& Makespan & 5046.9      & 4844.7     & 2233.4    & 2253.2         &{\ul2223.9}  & 2236.5           & 2114.5                      \\
                      &                                 & Gap      & 138.68\%    & 129.12\%   & 5.62\%    & 6.56\%         &{\ul5.17\%}  & 5.77\%            & -                         \\ \cline{2-10} 
                      & \multirow{2}{*}{$200 \times 5$} & Makespan & 9985.4      & 9674.7     & 5923.8    & 7093.4         &{\ul5915.1}  & 6069.1           & 5876.3                    \\
                      &                                 & Gap      & 69.93\%     & 64.64\%    & 0.81\%    & 20.71\%        &{\ul0.66\%}  & 3.28\%           & -                         \\ 
\hline\hline
\end{tabular}}
\end{table}

We further investigate the scalability and cross-size generalization of \textsc{DredRL} on the larger $30{\times}10$, $40{\times}10$, $100{\times}10$, and $200{\times}5$ instances. The model trained on small instances is directly transferred to these larger configurations without size-specific retraining. Notably, \textsc{DredRL} achieves the best learning-based makespan in all 8 large-scale settings when compared with the better of greedy and sampling decoding for DANIEL and HGNN. Relative to the best DANIEL result for each configuration, \textsc{DredRL} obtains an average makespan reduction of $1.23\%$, with improvements observed consistently across both SD1 and SD2. The gains are particularly evident on the largest SD1 instances. On SD1 $100{\times}10$, \textsc{DredRL} reduces the makespan from $946.0$ to $918.1$, corresponding to a $2.95\%$ improvement over DANIEL, while changing the OR-Tools-relative gap from $0.19\%$ to $-2.76\%$. An even larger improvement is obtained on SD1 $200{\times}5$, where \textsc{DredRL} reduces the best DANIEL makespan from $1901.1$ to $1833.9$, a $3.53\%$ reduction, and achieves a gap of $-2.69\%$ relative to OR-Tools. On SD2, \textsc{DredRL} also maintains consistent advantages as the problem scale increases. For example, on SD2 $30{\times}10$, it reduces the makespan from $757.48$ to $749.53$ and the corresponding gap from $9.42\%$ to $8.27\%$. On SD2 $40{\times}10$, \textsc{DredRL} further improves the best DANIEL makespan from $951.21$ to $943.43$, yielding the lowest gap of $-5.50\%$ among all compared methods. Even on the largest SD2 cases, the advantage is preserved: \textsc{DredRL} reduces the makespan from $2233.4$ to $2223.9$ on $100{\times}10$ and from $5923.8$ to $5915.1$ on $200{\times}5$. Interestingly, greedy decoding becomes preferable to sampling for \textsc{DredRL} on these two largest SD2 instances, suggesting that the relative benefit of stochastic decoding may depend on the problem scale. Overall, across all 16 configurations in Table~\ref{tab2:num_results}, \textsc{DredRL} outperforms the best DANIEL decoding in 13 cases and achieves an average makespan reduction of $1.34\%$. More importantly, its consistent advantage across all evaluated larger-scale instances demonstrates stronger cross-size generalization and scalability, suggesting that the learned operation--machine representations remain effective as the scheduling problem expands substantially beyond the training scale.

\subsubsection{Public Benchmarks} To verify the performance in solving real-world problems, we also evaluate \textsc{DredRL} based on a few commonly used public datasets \texttt{mk} and \texttt{la}, while \texttt{la} includes \texttt{rdata}, \texttt{edata}, \texttt{vdata}. All of the models are trained on the synthetic dataset which includes small $10 \times 5$ instances and tested on four public data benchmarks. Table~\ref{tab4:public} reports results on four public benchmarks and shows that \textsc{DredRL} achieves consistently strong performance against HGNN and DANIEL under both greedy and sampling decoding. As reported in Table~\ref{tab4:public}, on \texttt{mk}, \textsc{DredRL} reduces the makespan from $180.8$ with DANIEL to $178.1$, decreasing the relative gap to OR-Tools from $3.79\%$ to $2.24\%$. A similar advantage is observed on \texttt{edata}, where \textsc{DredRL} achieves a makespan of $1106.7$, compared with $1122.6$ for DANIEL, reducing the corresponding gap from $9.10\%$ to $7.56\%$. On \texttt{vdata}, \textsc{DredRL} further obtains the lowest learning-based makespan of $924.41$ and a gap of only $0.52\%$, slightly improving upon DANIEL at $925.4$ and $0.63\%$, respectively. The only exception is \texttt{rdata}, where DANIEL sampling achieves a marginally lower makespan of $978.28$ than \textsc{DredRL}'s $979.9$, corresponding to gaps of $4.54\%$ and $4.71\%$, respectively. Overall, \textsc{DredRL} achieves an average gap of $3.76\%$ to the time-limited OR-Tools reference across the four public benchmarks, compared with $4.52\%$ for DANIEL and $6.11\%$ for HGNN under their respective sampling modes. This corresponds to a $0.76$ percentage-point reduction in the average gap over DANIEL. These results demonstrate that \textsc{DredRL} learns transferable scheduling representations that generalize effectively beyond the training distribution.

\begin{table}[t]
\centering
\caption{Performance comparison of different algorithms on public benchmarks.}
\label{tab4:public}
\renewcommand{\arraystretch}{1.3}
\resizebox{\linewidth}{!}{
\begin{tabular}{ccc|cc|cc|cc|c}
\hline\hline
\multicolumn{3}{c|}{\multirow{2}{*}{Public Benchmarks}} & \multicolumn{2}{c|}{HGNN\cite{song2022flexible}} & \multicolumn{2}{c|}{DANIEL\cite{wang2023flexible}} & \multicolumn{2}{c|}{\textsc{DredRL} (Ours)} & \multirow{2}{*}{OR-Tools} \\
\multicolumn{3}{c|}{}                                                     & Greedy       & Sampling     & Greedy      & Sampling       &   Greedy      & Sampling           &                           \\ \hline
\multicolumn{2}{c}{\multirow{2}{*}{\texttt{mk}}}               & Makespan & 201.4        & 190.3        & 185.7       & 180.8           & 182.8         & {\ul 178.1}      & 174.2                     \\
\multicolumn{2}{c}{}                                           & Gap      & 15.61\%      & 9.41\%       & 6.60\%      & 3.79\%          & 4.94\%        & {\ul 2.24\%}     & -                         \\ \hline
\multirow{6}{*}{\texttt{la}} & \multirow{2}{*}{\texttt{rdata}} & Makespan & 1030.83      & 985.3        & 1031.63     & {\ul 978.28}    & 1026.6        & 979.9            & 935.8                     \\
                    &                                          & Gap      & 10.15\%      & 5.29\%       & 10.24\%     & {\ul 4.54\%}    & 9.70\%        & 4.71\%           & -                         \\ \cline{2-10} 
                    & \multirow{2}{*}{\texttt{edata}}          & Makespan & 1187.48      & 1116.68      & 1194.98     & 1122.6          & 1164.8        & {\ul 1106.7}     & 1028.9                   \\
                    &                                          & Gap      & 15.41\%      & 8.53\%       & 16.14\%     & 9.10\%          & 13.21\%       & {\ul 7.56\%}     & -                         \\ \cline{2-10} 
                    & \multirow{2}{*}{\texttt{vdata}}          & Makespan & 955.9        & 930.80       & 944.85      & 925.4           & 941.7         & {\ul 924.41}     & 919.6                     \\
                    &                                          & Gap      & 3.95\%       & 1.22\%       & 2.75\%      & 0.63\%          & 2.40\%       & {\ul 0.52\%}     & -                         \\ 
\hline\hline
\end{tabular}}
\end{table}

\vspace{-8pt}

\subsection{Stochastic Research}

To further evaluate \textsc{DredRL} under stochastic processing times, we conduct experiments on the SD3 dataset following the generation protocol in~\cite{smit2025neural}. SD3 preserves the instance structure of SD2 but introduces stronger correlation among the processing times of an operation across compatible machines. Specifically, a base processing time $p_{ij}$ is first sampled from $\mathcal{U}(1,99)$ for each operation $O_{ij}$, and its machine-dependent processing time is then generated as $p_{ij}^{k}\sim\mathcal{U}(0.85p_{ij},1.15p_{ij})$. To construct stochastic instances, the deterministic processing time is treated as the median, while the coefficient of variation is sampled from $\mathcal{U}(0.1,0.5)$ to determine the uncertainty level of each operation--machine pair. Following the evaluation setting in~\cite{smit2025neural}, we consider four instance sizes, $10{\times}5$, $20{\times}5$, $15{\times}10$, and $20{\times}10$, and compare \textsc{DredRL} with DANIEL, SPM-DAN, and the OR-Tools' CP-SAT solver and CP-Stoch algorithm\cite{smit2025neural} baselines. Both greedy and sampling inference are reported for the learning-based methods to provide a consistent comparison under stochastic scheduling conditions.

As shown in Table~\ref{tab5:num_results_stoch}, \textsc{DredRL} demonstrates strong performance on SD3, particularly under sampling inference. Compared with DANIEL, \textsc{DredRL} achieves a lower sampling makespan on all four instance sizes, with reductions of $1.74\%$, $2.81\%$, $2.53\%$, and $3.06\%$ for $10{\times}5$, $20{\times}5$, $15{\times}10$, and $20{\times}10$, respectively. More importantly, \textsc{DredRL} achieves the best learning-based results on three of the four configurations, outperforming the stochasticity-aware SPM-DAN on $10{\times}5$, $20{\times}5$, and $15{\times}10$. Its gaps to the deterministic OR-Tools reference are consistently negative under sampling, ranging from $-1.39\%$ to $-4.11\%$, indicating that \textsc{DredRL} obtains better schedules than the time-limited OR-Tools reference on all settings. It also outperforms CP-Stoch on three of the four problem sizes, with only a marginal difference on $10{\times}5$. Although SPM-DAN achieves the best result on $20{\times}10$, the consistently competitive performance of \textsc{DredRL} across different problem sizes demonstrates its ability to capture processing-time uncertainty without relying on a dedicated scenario-processing architecture. This highlights the robustness and adaptability of the representation and diffusion mechanism.

\begin{table}[t]
\centering
\caption{Performance comparison of different algorithms on synthetic datasets SD3 with Different Instance Sizes.}
\label{tab5:num_results_stoch}
\renewcommand{\arraystretch}{1.3}
\resizebox{\linewidth}{!}{
\begin{tabular}{ccc|cc|cc|cc|cc}
\hline\hline
\multicolumn{3}{c|}{\multirow{2}{*}{Size}}                         & \multicolumn{2}{c|}{DANIEL\cite{wang2023flexible}}& \multicolumn{2}{c|}{SPM-DAN\cite{smit2025neural}} & \multicolumn{2}{c|}{\textsc{DredRL} (Ours)} & \multicolumn{2}{c}{OR-Tools} \\
\multicolumn{3}{c|}{}                                              & Greedy   & Sampling        & Greedy     & Sampling        & Greedy         & Sampling            & CP-SAT     & CP-Stoch\cite{smit2025neural}     \\ \hline
\multirow{8}{*}{SD3} & \multirow{2}{*}{$10 \times 5$}   & Makespan & 722.77   & 688.74          & 710.63     & 677.24          & 712.30    &    {\ul676.74}          & 688.10     & 676.19       \\
                     &                                  & Gap      & 5.04\%   & 0.09\%          & 3.27\%     & -1.58\%         & 3.52\%    &    {\ul-1.65\%}         & -          & -1.73\%      \\ \cline{2-11} 
                     & \multirow{2}{*}{$20 \times 5$}   & Makespan & 1307.48  & 1272.08         & 1279.70    & 1243.75         & 1276.49   &    {\ul1236.30}         & 1253.67    & 1271.71      \\
                     &                                  & Gap      & 4.29\%   & 1.47\%          & 2.08\%     & -0.79\%         & 1.82\%    &    {\ul-1.39\%}         & -          & 1.44\%       \\ \cline{2-11} 
                     & \multirow{2}{*}{$15 \times 10$}  & Makespan & 1111.55  & 1066.60         & 1090.92    & 1039.73         & 1091.03   &    {\ul1039.59}         & 1074.88    & 1105.90      \\
                     &                                  & Gap      & 3.41\%   & -0.77\%         & 1.49\%     & -3.27\%         & 1.50\%    &    {\ul-3.28\%}         & -          & 2.89\%       \\ \cline{2-11} 
                     & \multirow{2}{*}{$20 \times 10$}  & Makespan & 1376.18  & 1336.97         & 1289.77    & {\ul1261.80}         & 1302.80   &    1296.06          & 1351.58    & 1406.68      \\
                     &                                  & Gap      & 1.82\%   & -1.08\%         & -4.57\%    & {\ul-6.64\%}         & -3.61\%   &    -4.11\%          & -          & 4.08\%       \\ 
                     \hline\hline
\end{tabular}}
\end{table}

\vspace{-8pt}

\subsection{Ablation Study on Representation and Diffusion}

To investigate whether the performance gain comes from the proposed \textsc{DredRL} representation, the diffusion-based decision model, or their combination, we organize a controlled comparison with four configurations: DANIEL, \textsc{DredRL}-MLP, proposed \textsc{DredRL} (we note as \textsc{DredRL}-Diffusion below), and DANIEL-Diffusion, all using greedy method. This design allows us to separately examine the effect of replacing the original representation with \textsc{DredRL}, introducing diffusion into different model architectures, and the interaction between the two components. We evaluate all configurations on six SD2 problem sizes. For each setting, we report the mean and standard deviation of makespan over five training seeds to assess both solution quality and training stability. We then compare the relative gap to the OR-Tools reference across test instances, where proven optimal solutions are used when available and the time-limited OR-Tools solutions are used as references otherwise. 

\begin{figure}[h!]
\includegraphics[width=0.5\textwidth]{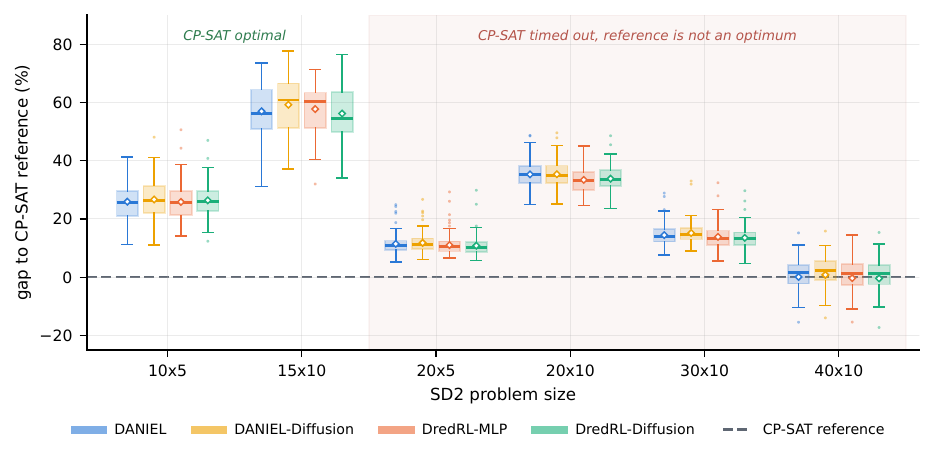}
\caption{Comparison of relative makespan gaps to the OR-Tools reference on SD2 across different problem sizes.}
\label{img3:fin_gap_vs_ortools}
\vspace{-5pt}
\end{figure}

The results in Figure~\ref{img3:fin_gap_vs_ortools} demonstrate the advantage of incorporating diffusion into \textsc{DredRL}. Among the four methods, \textsc{DredRL}-Diffusion achieves the best mean makespan on four of the six problem sizes ($15{\times}10$, $20{\times}5$, $30{\times}10$, and $40{\times}10$), while \textsc{DredRL}-MLP performs best on the remaining two. In terms of solution quality relative to OR-Tools, \textsc{DredRL}-Diffusion also obtains the lowest average gap of $23.36\%$ across all problem sizes, compared with $23.56\%$ for \textsc{DredRL}-MLP and $24.00\%$ for DANIEL. Notably, it achieves increasingly competitive results on larger instances, including gaps of $13.48\%$ and $-0.44\%$ on $30{\times}10$ and $40{\times}10$, respectively. In addition to solution quality, \textsc{DredRL}-Diffusion exhibits competitive stability and robustness across the five random seeds, with an average makespan standard deviation of $4.04$, compared with $4.08$ for \textsc{DredRL}-MLP and $4.73$ for DANIEL. In contrast, directly combining diffusion with DANIEL results in substantially higher training variance, with an average standard deviation of $14.74$. Even on the larger $30{\times}10$ and $40{\times}10$ instances, where the variance of \textsc{DredRL}-Diffusion increases to $7.31$ and $7.34$, it remains substantially below the $20.24$ and $22.81$ observed for DANIEL-Diffusion. These results suggest that \textsc{DredRL}-Diffusion provides the strongest overall trade-off between solution quality, scalability, and training stability, and is therefore adopted as the final configuration.

\vspace{-8pt}

\section{Conclusion}

This paper presented \textsc{DredRL}, a dual-representation-enhanced diffusion-policy reinforcement learning framework for FJSP. By coupling operation and machine representations through enhanced self-attention and cross-attention, \textsc{DredRL} captures both intra-entity dependencies and inter-entity compatibility more effectively than prior DRL schedulers. In addition, a diffusion model is employed as the policy network to improve the expressiveness and robustness of scheduling decisions. Extensive experiments on synthetic, public, large-scale, and stochastic FJSP benchmarks demonstrate that \textsc{DredRL} consistently outperforms SOTA learning-based methods across most configurations. Compared with DANIEL, it reduces makespan by up to $5.44\%$ on small-scale instances and $3.53\%$ on larger-scale instances without size-specific retraining. It also achieves the best learning-based performance on most public benchmarks and outperforms the stochasticity-aware SPM-DAN on three of four SD3 configurations. These results demonstrate strong cross-size and cross-distribution generalization, and robustness to processing-time uncertainty, showing the effectiveness of combining enhanced dual representations with diffusion-based policies for scalable FJSP scheduling.

Future work will extend \textsc{DredRL} to multi-objective optimization and complex dynamic environments, including machine breakdowns and new job arrivals. We will also investigate more efficient diffusion-policy inference, such as reducing denoising steps and policy distillation, to further improve computational efficiency and facilitate industrial deployment.

\vspace{-5pt}

\bibliographystyle{IEEEtran}
\bibliography{IEEEabrv,references}
 




\vfill

\end{document}